%% file: main.tex
\documentclass[letterpaper]{article} 
\usepackage{aaai25}  
\usepackage{times}  
\usepackage{helvet}  
\usepackage{courier}  
\usepackage[hyphens]{url}  
\usepackage{graphicx} 
\usepackage{natbib}  
\usepackage{caption} 
\usepackage{newfloat}
\usepackage{listings}

\DeclareCaptionStyle{ruled}{labelfont=normalfont,labelsep=colon,strut=off} 
\usepackage{moresize}
\usepackage{amsmath}
\usepackage{algorithmic}
\usepackage{comment}
\usepackage{paralist}
\usepackage{bm}
\usepackage{pgfplots}
\usetikzlibrary{pgfplots.dateplot}
\usepackage[english]{babel}
\usepackage[latin1]{inputenc}
\usepackage{mathrsfs}
\usepackage{graphicx}

\usepackage{amssymb}
\usepackage{amsfonts}
\usepackage{longtable}
\usepackage{rotating}
\usepackage{multirow}
\usepackage{subfigure}
\usepackage{enumitem}
\usepackage[linesnumbered,algoruled,boxed,lined]{algorithm2e}
\usepackage{adjustbox}
\usepackage{threeparttable}
\usepackage{booktabs}
\usepackage{colortbl}
\usepackage{marginnote}

\definecolor{tblue}{RGB}{51,139,180}
\definecolor{torange}{RGB}{235,157,14}
\definecolor{tgreen}{RGB}{44,170,44}
\definecolor{tred}{RGB}{214,99,40}
\definecolor{tpurple}{RGB}{148,133,189}
\definecolor{LightCyan}{rgb}{0.88,1,1}
\definecolor{hidden-draw}{RGB}{40,158,106}
\definecolor{hidden-pink}{RGB}{255,245,247}
\definecolor{lightred}{RGB}{255, 204, 204}
\definecolor{lightgreen}{RGB}{224, 255, 225}
\definecolor{lightyellow}{RGB}{255, 241, 224}
\definecolor{lightblue}{rgb}{0.901, 1, 1}
\definecolor{lightpurple}{RGB}{225, 225, 255}
\definecolor{purple}{RGB}{202, 206, 218}
\definecolor{lightgray}{gray}{0.9}

\newcommand{\ie}{\textit{i}.\textit{e}.}

\newcommand{\qr}[1]{{{\textcolor{red}{[Qianru: #1]}}}}

\renewcommand{\qr}{}

\def\model{LightST}

\begin{document}

\title{Efficient Traffic Prediction Through Spatio-Temporal Distillation}
\author{
    Qianru Zhang \textsuperscript{\rm 1},
    Xinyi Gao \textsuperscript{\rm 2},
    Haixin Wang \textsuperscript{\rm 3},
    Siu-Ming Yiu \textsuperscript{\rm 1}\thanks{Corresponding Author},
    Hongzhi Yin \textsuperscript{\rm 2}\thanks{Corresponding Author}\\
}

\affiliations{
    \textsuperscript{\rm 1} The University of Hong Kong\\
    \textsuperscript{\rm 2} The University of Queensland\\
    \textsuperscript{\rm 3} University of California, Los Angles\\
}

\maketitle

\begin{abstract}
Graph neural networks (GNNs) have gained considerable attention in recent years for traffic flow prediction due to their ability to learn spatio-temporal pattern representations through a graph-based message-passing framework. Although GNNs have shown great promise in handling traffic datasets, their deployment in real-life applications has been hindered by scalability constraints arising from high-order message passing. Additionally, the over-smoothing problem of GNNs may lead to indistinguishable region representations as the number of layers increases, resulting in performance degradation. To address these challenges, we propose a new knowledge distillation paradigm termed LightST that transfers spatial and temporal knowledge from a high-capacity teacher to a lightweight student. Specifically, we introduce a spatio-temporal knowledge distillation framework that helps student MLPs capture graph-structured global spatio-temporal patterns while alleviating the over-smoothing effect with adaptive knowledge distillation. 
Extensive experiments verify that LightST significantly speeds up traffic flow predictions by 5X to 40X compared to state-of-the-art spatio-temporal GNNs, all while maintaining superior accuracy.
\end{abstract}

\input{intro}

\input{solution2}

\input{eval}

\input{relate}

\input{conclusion}

\appendix
\input{appendix}

\bibliography{aaai25}

\end{document}

%% file: intro.tex
\section{Introduction}
\label{sec:intro}

Recent advancements in intelligent transportation systems have seen significant progress in traffic flow prediction~\cite{wang2020traffic,pan2019urban,liang2019urbanfm} through the development of Graph Neural Networks (GNNs). GNN-based methods~\cite{gao2024graph1,gao2024graph2,gao2024graphsurvey,zhang2024survey1} utilize the message-passing mechanism to propagate embeddings, enabling them to capture spatio-temporal traffic patterns. For example, STGCN~\cite{STGCN}, AGCRN~\cite{AGCRN} and GWN~\cite{shleifer2019incrementally} are built upon GNNs for traffic prediction~\cite{zhang2024survey}. GCN-based models~\cite{STGCN,shleifer2019incrementally} use convolutional operations on the graph structure to extract spatial features of traffic data, while GAT-based methods~\cite{han2022lst} employ attention mechanisms to weigh the importance of graph-based neighboring locations for each region, propagating information.

The effectiveness of spatio-temporal GNNs is largely attributed to the complex model structure and recursive message-passing architecture that encodes high-order region-wise connectives and learns region representations. However, the increasing complexity of larger and deeper GNN model structures leads to the computationally intensive \qr{inference procedure}, posing challenges for practical applications due to the scalability constraints. Therefore, a lightweight yet effective traffic prediction framework is required for practical settings of intelligent transportation systems. Additionally, propagating embeddings across multiple layers gradually makes node features more uniform, ultimately reducing the model's ability to differentiate node features~\cite{chen2020simple,zhou2020towards}. This inherent recursive message-passing paradigm may fall short in encoding diverse spatio-temporal patterns, ultimately degrading traffic prediction performance.

To mitigate these challenges, existing methods~\cite{izadi2024knowledge,wang2024spatial,zhang2021graph} leverage knowledge distillation (KD) to transfer knowledge from complex teacher models to smaller student models, helping to overcome limitations in spatio-temporal graph neural networks (ST-GNNs). However, they still face challenges and achieve sub-optimal performances. KD-pruning method~\cite{izadi2024knowledge} calculates pruning scores via cost function and fine-tunes the student network decomposed with GNNs, but they do not fully address the over-smoothing issue inherent to GNNs. Furthermore, incorporating GNNs into the student network does not effectively resolve the high computational cost associated with training and deploying these models. A recent work ~\cite{wang2024spatial} proposes a Spatial-Temporal Knowledge Distillation (STKD) algorithm framework for lightweight network traffic anomaly detection, integrating multi-scale 1D CNNs and LSTMs with identity mapping for performance enhancement. However, the individual components of STKD, namely 1D CNNs and LSTMs, may not be optimal for capturing temporal correlations, potentially limiting its effectiveness in spatio-temporal domains. Firstly, 1D CNNs function within fixed-size windows, potentially impeding their ability to capture essential long-range dependencies critical for modeling intricate temporal patterns within traffic data. Secondly, both 1D CNNs and LSTMs generate fixed-length representations, potentially compromising detailed temporal nuances essential for capturing subtle variations and complex temporal relationships in dynamic traffic environments.

\begin{figure}[t]
\centering
  \begin{minipage}{0.45\textwidth}
	\includegraphics[width=\textwidth]{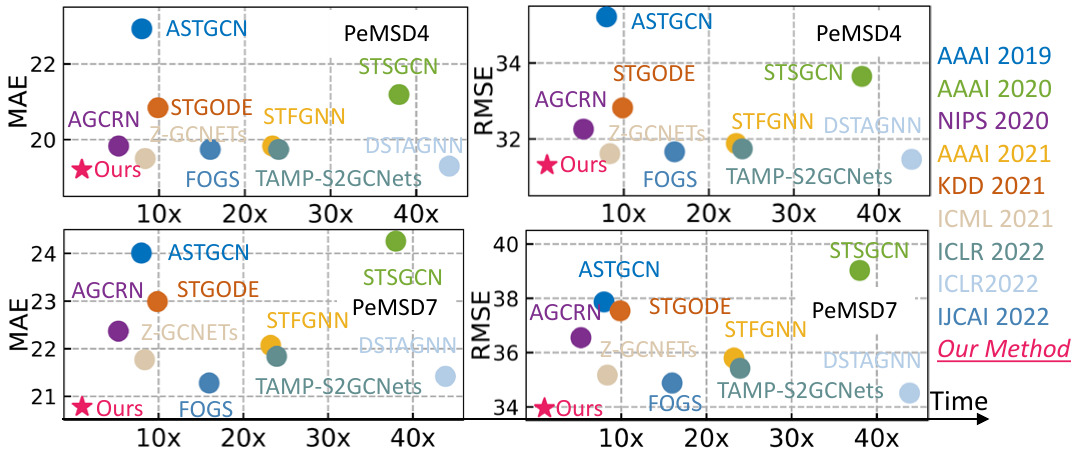}
  \end{minipage}
\caption{Model performance comparison in terms of both traffic flow prediction accuracy and inference time. Lower MAE and RMSE indicate better performance. The symbol $40\times$ indicates that our \model\ runs 40 times faster than a reference baseline method measured by inference time.}\label{fig:intro}
\end{figure}

In light of the motivations described above, this study aims to: i) develop a traffic prediction model that is highly scalable while effectively capturing complex spatio-temporal dependency patterns across various locations and time periods; and ii) enhance the spatio-temporal encoding function to mitigate the issue of over-smoothing. To achieve this, we propose a dual-level spatio-temporal knowledge distillation paradigm that effectively transfers complex dynamic spatio-temporal dependency knowledge into a compact and fast-to-execute student model. Specifically, in the distillation procedure, the soft prediction labels from the teacher GNN guide the learning of the student model. This transfer of knowledge effectively incorporates structural information from both spatial and temporal aspects. Furthermore, to avoid using a uniform alignment factor for all region pairs, we propose an adaptive embedding-level distillation framework that enhances the knowledge transfer while mitigating over-smoothing effects. Our empirical studies, as shown in Figure~\ref{fig:intro}, suggest that our designed spatio-temporal knowledge distillation substantially benefits traffic prediction performance in terms of both forecasting accuracy and efficiency. Our key contributions are listed as follows:
\begin{itemize}[leftmargin=*]

\item {To overcome the computational and oversmoothing challenges in state-of-the-art GNN-based traffic prediction models, we propose distilling the complex spatio-temporal GNN architecture into a streamlined MLP model, enhancing both efficiency and robustness in traffic prediction.}

\item We design a new spatio-temporal knowledge distillation paradigm with two model alignment levels, enabling the transfer of knowledge related to spatial and temporal dynamics while enhancing the student model with a global context. Additionally, we propose an adaptive contrastive distillation scheme to further enhance the robustness of the prediction model against the over-smoothing issue.

\item We empirically validate our new framework on 5 real-world traffic datasets. Evaluation results demonstrate that our model achieves state-of-the-art traffic prediction accuracy,  with a 5$\times$ to 40$\times$ inference speedup compared to existing baselines. Our codes are available at: \url{https://github.com/lizzyhku/TP/tree/main}. We also attached the reproducibility checklist.

\end{itemize}

%% file: solution2.tex
\section{Methodology}
\label{sec:solution}

\subsection{Spatio-temporal Graph and Traffic Data}

Following established practices~\cite{lan2022dstagnn,chen2021z}, we define our spatial graph $\mathcal{G} = (\mathcal{V}, \mathcal{E})$ to represent the geographically-adjacent relationships between traffic sensing regions. We represent the traffic volume data across both spatial and temporal dimensions using a matrix $\textbf{X} \in \mathbb{R}^{N \times T}$, where $N$ denotes the number of regions and $T$ represents the number of time slots. Each element $x_{n,t}$ within $\textbf{X}$ corresponds to the traffic volume information of region $n$ during time slot $t$.

\noindent\textbf{Problem Formulation}. Our goal is to develop a function $\mathcal{F}$ that predicts future traffic flow $\mathbf{\hat{Y}} \in \mathbb{R}^{N \times H}$ based on observed traffic flow data $\textbf{X} = ( x_1, ..., x_T) \in \mathbb{R}^{N \times T}$. The ground truth is $\mathbf{Y} = (x_{T+1}, ..., x_{T+H}) \in \mathbb{R}^{N \times H}$. Here, $\textbf{X}$ represents the observed traffic flow from $N$ sensing regions within a sensor graph $\mathcal{G}$ over the preceding $T$ time slots, and $\mathbf{\hat{Y}}$ denotes the predicted traffic flow for the subsequent $H$ time steps.  Our approach aims to learn $\mathcal{F}$ while preserving both spatial and temporal dependencies within the data. This is shown as $\mathbf{\hat{Y}} = \mathcal{F}(\textbf{X}; \mathcal{G})$.

\subsection{Spatio-Temporal Graph Neural Networks}
\noindent\textbf{Data Scale}. Graph Neural Networks (GNNs) have demonstrated significant potential in learning from spatio-temporal data~\cite{DCRNN,lan2022dstagnn,rao2022fogs}. Inspired by this, our teacher model leverages a graph-based message passing framework that harnesses the power of GNNs to capture region-wise dependencies. We begin by mapping traffic data into a latent representation space. Each element ${\mathbf{X}}_{n,t} \in \textbf{X}$ is encoded into an embedding $\mathbf{E}^{(\text{d})}_{n,t} \in \mathbb{R}^d$ as follows:

\begin{align}
\mathbf{E}^{(\text{d})}_{n,t} = \text{Z-Score}(x_{n,t}) \cdot \mathbf{e} = \frac{x_{n,t}-\mu}{\sigma}\cdot \mathbf{e}.
\end{align}

We normalize the value of ${\mathbf{X}}_{n,t}$ using the Z-Score function, which centers the data around zero with a standard deviation of one. This normalized value is then multiplied by a base embedding vector $\mathbf{e} \in \mathbb{R}^d$, resulting in the final embedding $\mathbf{E}^{(\text{d})}_{n,t} \in \mathbb{R}^d$. $\mu$ represents the average traffic flow value of node $n$ over the last 12 time steps. And $\sigma$ denotes the standard deviation of the traffic flow values of node $n$ over the previous 12 time steps. The base embedding vector $\mathbf{e}$ acts as a template, and the Z-Score normalization scales the traffic flow value, effectively creating a unique embedding for each data point based on its relative position within the normalized distribution.

\begin{figure*}[t]
\centering
\hspace{-5.5mm}
  \begin{minipage}{0.85\linewidth}
	\includegraphics[width=\linewidth]{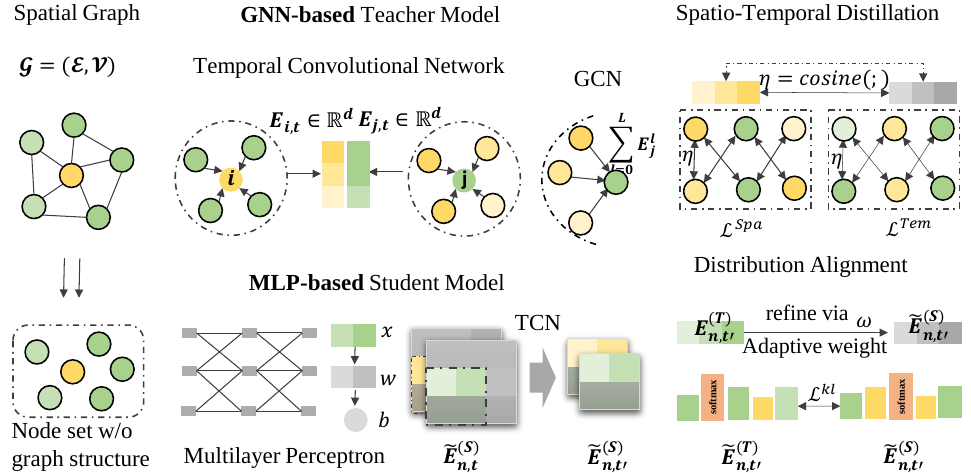}
  \end{minipage}\hspace{-3.0mm}
\caption{Our proposed spatio-temporal knowledge distillation framework, \model, comprises two main components: a GNN-based teacher model and an MLP-based student model. The distillation paradigm itself is structured in two parts: spatio-temporal distillation and distribution alignment.}
\vspace*{-4mm}
\label{fig:framework}
\end{figure*}

\noindent\textbf{Time-aware Spatial Message Passing}.
To capture time-aware spatial dependencies and learn representations specific to each time slot $t$, we employ a time-specific message passing mechanism among traffic sensing regions. This process involves aggregating information from neighboring nodes within the sensor graph. The embedding of region $n$ at the $l$-th GNN layer is denoted by $\mathbf{E}_{n}^{(l)}$, and is shown as follows:

\begin{equation}
\begin{aligned}
    \mathbf{E}_{n}^{(l)} &= \sigma^{(1)}(\sum_{j\in {\mathcal{N}_n}} \alpha_{n,j} \mathbf{W}^{(l-1)} \mathbf{E}_{j}^{(l-1)}),\\
    \alpha_{n,j} &= \frac{1}{\sqrt{|\mathcal{N}_n||\mathcal{N}_j|}}.
\end{aligned}
\end{equation}

Here, $\mathcal{N}_n$ and $\mathcal{N}_j$ represent the sets of neighboring nodes for regions $n$ and $j$, respectively. The embedding propagation mechanism aggregates information from these neighboring nodes. $\sigma^{(1)}(\cdot)$ represents the ReLU activation function. The normalization weight $\alpha_{n,j}\in\mathbb{R}$ for a node pair $(n, j)$ is calculated based on the degrees of the nodes. This cross-layer information propagation and aggregation can be formalized in matrix form using the adjacency matrix $\mathbf{A}\in\mathbb{R}^{N\times N}$ of our spatial graph $\mathcal{G}$:

\begin{align}
    \label{eq:spress}
    \mathbf{E}^{(l)} = \sigma^{(1)}(\mathbf{D}^{-\frac{1}{2}} \mathbf{A} \mathbf{D}^{-\frac{1}{2}} \mathbf{E}^{(l-1)}\mathbf{W}^{(l-1)\top}).
\end{align}

The embedding matrix $\mathbf{E} = \sum_{l=0}^L \mathbf{E}^{(l)} \in\mathbb{R}^{N\times d}$ contains node embeddings, where each row represents individual traffic sensing point $\mathbf{E}_{n}$ $(1 \le n \le N)$. $\mathbf{D}\in\mathbb{R}^{N\times N}$ denotes the diagonal degree matrix, and $L$ represents the number of graph neural iterations. The final aggregated region representations are given by $\mathbf{E} \in\mathbb{R}^{N\times d}$.

\noindent\textbf{Temporal Encoder Layer}. To capture temporal dependencies across all sensor regions, we employ a two-layer Temporal Convolutional Network (TCN) to model temporal correlations. This can be represented as follows:
\begin{align}
\label{eq:tcn}
\tilde{\mathbf{E}}_{n} = \sigma^{(2)}(\delta(\mathbf{W}\ast \mathbf{E}_{n}+\mathbf{b})+ \mathbf{E}_{t}), t \in [1,T].
\end{align}
Here, the embedding matrix of all $N$ regions at the time slot of $t$ is represented by $\mathbf{E}_{t} \in \mathbb{R}^{N \times d}$. For a specific region $n$, the embedding matrix is represented by $\mathbf{E}_{n} \in \mathbb{R}^{T \times d}$, which captures the embedding of region $n$ across the previous $T$ time slots. The temporal convolution kernel and bias are represented by $\mathbf{W} \in \mathbb{R}^{f \times d}$ and $\textbf{b} \in \mathbb{R}^{d}$, respectively. These are learnable transformation parameters. Here, $f$ denotes the kernel size. The operation $\ast$ denotes the temporal convolution, while the dropout $\delta(\cdot)$ and LeakyReLU activation function $\sigma^{(2)}(\cdot)$ are used as well. The output representation $\tilde{\mathbf{E}} \in \mathbb{R}^{N \times T  \times d}$ is generated with two-layer TCNs that serve as the temporal encoder in the teacher model, so as to capture the time-evolving traffic patterns.

\subsection{Distillation Process}
This research aims to develop a \emph{lightweight} and \emph{efficient} traffic predictor that leverages spatial and temporal knowledge extracted from a Graph Neural Network (GNN) teacher model to achieve accurate traffic predictions. This is achieved through both explicit spatio-temporal distillation and implicit knowledge transfer from the GNN teacher model to an MLP-based student model via distribution alignment. We provide details of the distillation process as follows:

\noindent \textbf{Spatio-Temporal Distillation}.
During this stage, we aim to align the probability distributions of the teacher and student models, thereby enhancing the predictive performance of the student model. This is achieved through knowledge distillation, where the output results of the teacher model are used as soft labels to guide the student model. We minimize the difference between the probability distributions of the teacher and student models by employing an MSE-based loss function~\cite{wu2020connecting}. The MSE-based loss functions for the teacher and student models in traffic prediction are defined as follows:
\begin{equation}
\begin{aligned}
    \label{eq:lossgcn}
    \mathcal{L}^{(\text{T})} &= \frac{1}{N} \sum\limits_{n = 1}^{N}\sum \limits_{t'=T+1}^{T+H}(\hat{\textbf{Y}}^{(\text{T})}_{{n},t'}-\textbf{Y}_{n,t'})^2, \\
    \mathcal{L}^{(\text{S})} &= \frac{1}{N} \sum\limits_{i = 1}^{N}\sum \limits_{t'=T+1}^{T+H}(\hat{\textbf{Y}}^{(\text{S})}_{{n},t'}-\textbf{Y}_{n,t'})^2.
\end{aligned}
\end{equation}

\noindent Here, $\mathcal{L}^{(\text{T})}$ represents the loss of the GNN-based teacher model, denoted by the superscript $(\text{T})$. Similarly, $\mathcal{L}^{(\text{S})}$ represents the loss of the MLP-based student model, denoted by the superscript $(\text{S})$. $N$ corresponds to the number of sensing regions within the spatio-temporal graph $\mathcal{G}$, and $H$ denotes the number of predicted time steps. $\hat{\textbf{Y}}^{(\text{T})}_{{n},t'}$ represents the predicted traffic volume of node $n$ at time step $t'$ by the teacher model, while $\hat{\textbf{Y}}^{(\text{S})}_{{n},t'}$ represents the predicted traffic volume of the same node and time step by the student model. $\textbf{Y}_{n,t'}$ denotes the ground truth traffic volume of node $n$ at time step $t'$. To further measure the distribution difference of the GNN-based teacher model and that of the MLP-based student model, we adopt Kullback Leibler (KL)~\cite{hu2013kullback} divergence into the distribution alignment loss $\mathcal{L}^{(\text{KL})}$:

\begin{align}
    \label{eq:losspredis}
    \mathcal{L}^{(\text{KL})} = \sum_{n = 1}^{N} \sum_{t' = T+1}^{T+H} \tilde{\textbf{E}}_{n,t'}^{(\text{T})} \cdot \log(\text{Softmax}(\tilde{\textbf{E}}_{n,t'}^{(\text{S})})).
\end{align}

\noindent Here, $\mathcal{L}^{(\text{KL})}$ represents the Kullback-Leibler (KL) divergence loss, denoted by the superscript $(\text{KL})$. $\tilde{\textbf{E}}_{n,t'}^{(\text{T})}$ denotes the output of the Temporal Convolutional Network (TCN) for node $n$ at time step $t'$ from the teacher model, while $\tilde{\textbf{E}}_{n,t'}^{(\text{S})}$ represents the corresponding output from the student model. Minimizing $\mathcal{L}^{(\text{KL})}$ aims to align the predicted probability distribution of the student model with that of the teacher model. This process effectively incorporates spatio-temporal knowledge from the teacher into the student's predictions.

\noindent \textbf{Distribution Alignment}. 
To further transfer spatio-temporal knowledge from the representation space, we introduce distribution alignment. Our model aims to achieve consistency in the embeddings of the same region from both the teacher and student models. To account for variations in the spatio-temporal context within the latent semantic space, we assign different consistency strengths to embeddings of different region pairs. This is achieved using a similarity function denoted by $\eta(\cdot)$, where we leverage embedding cosine similarity. 

Our approach, which involves distillation from both the spatial (with loss $\mathcal{L}^{(\text{P})}$) and temporal (with loss $\mathcal{L}^{(\text{E})}$) dimensions using contrastive learning, is formally presented as follows:

\begin{equation}
\begin{aligned}
    \label{eq:loss_sp}
    \mathcal{L}^{(\text{P})} &= \sum \limits_{n = 1}^{{N}} \sum_{t' = T+1}^{T+H}-\log \frac{\text{exp}(\frac{\eta(\tilde{\textbf{E}}^{(\text{S})}_{n,t'},\textbf{E}^{(\text{T})}_{n,t'})}{\tau_{2}})}{\sum \limits_{n' \neq n}\text{exp}(\frac{\eta(\tilde{\textbf{E}}^{(\text{S})}_{n',t'},\textbf{E}^{(\text{T})}_{n,t'})}{\tau_{2}})}, \\
    \mathcal{L}^{(\text{E})} &= \sum \limits_{n=1}^{N} \sum_{t' = T+1}^{T+H} -\log \frac{\text{exp}(\frac{\eta(\tilde{\textbf{E}}^{(\text{S})}_{n,t'},\tilde{\textbf{E}}^{(\text{T})}_{n,t'})}{\tau_{3}})}{\sum \limits_{n' \neq n}\text{exp}(\frac{\eta(\tilde{\textbf{E}}^{(\text{S})}_{n,t'},\tilde{\textbf{E}}^{(\text{T})}_{n,t'})}{\tau_{3}})}.
\end{aligned}
\end{equation}
\noindent The parameters $\tau_{2}$ and $\tau_{3}$ control the temperature of the softmax function used in the contrastive loss during training. $\mathcal{L}^{(\text{P})}$ and $\mathcal{L}^{(\text{E})}$ represent the spatial loss and temporal loss, respectively, denoted by the superscripts $(\text{P})$ and $(\text{E})$. $\textbf{E}^{(\text{T})}_{n,t'}$ represents the output embedding from the GCN layers of the teacher model for region $n$ at time step $t'$. Similarly, $\tilde{\textbf{E}}^{(\text{T})}_{n,t'}$ represents the output embedding from the TCN layers of the teacher model for the same region and time step. Finally, $\tilde{\textbf{E}}^{(\text{S})}_{n,t'}$ denotes the output embedding from TCN layers of the student model for region $n$ at time step $t'$.

\noindent \textbf{Model Optimization}. 
Following the learning paradigm of knowledge distillation, we first train the GNN-based teacher model of \model\ until convergence using the loss function $\mathcal{L}^{(\text{T})}$ from Equation~\ref{eq:lossgcn}. This involves feeding mini-batches of traffic observation tensors into the model and optimizing it. We then perform joint training to optimize both the MLP-based student model and the teacher model together. The overall objective function is an integration of the optimized objectives, which is shown as follows:
\begin{align}
    \label{eq:loss_sum}
    \mathcal{L} = \mathcal{L}^{(\text{S})} + \lambda_1 \cdot \mathcal{L}^{(\text{KL})} + \lambda_2 \cdot (\mathcal{L}^{(\text{P})} + \mathcal{L}^{(\text{E})}),
\end{align}
where \noindent $\lambda_1$ and $\lambda_2$ are loss weights. The training process of our \model\ is elaborated in Algorithm 1 in Appendix A.1.

\subsection{Discussion of Model}

\noindent \textbf{Model Complexity}.
To evaluate the efficiency improvement of the proposed MLP-based student model, we analyzed its time complexity in comparison to the GNN-based teacher model. The teacher model has a higher time complexity due to its graph information propagation in the encoder, which requires $\mathcal{O}(|\mathcal{E}|\times L\times d)$, where $|\mathcal{E}|$ is the number of edges, and $L$ is the number of graph layers. In contrast, the proposed student model only requires $\mathcal{O}(B\times L'\times d^2)$ for CL-enhanced distribution alignment, where $B$ is the number of samples in each batch, and $L'$ is the number of MLP layers. In summary, our analysis shows that the proposed MLP-based student model is significantly more efficient than the GNN-based traffic prediction methods, making it a promising framework for large-scale traffic data in practical spatio-temporal data mining scenarios.

\noindent \textbf{Model Theoretical Analysis}. We discuss how adaptive spatio-temporal distillation can alleviate the over-smoothing effects in spatio-temporal GNNs. We will begin by introducing the message passing schema that is used to propagate information along the graph-structured path in our spatial graph $\mathcal{G}$:
\begin{align}
    \label{eq:devitation}
    \mathbf{E}_{n}^{(\text{T})(L)}  = \sum \limits_{j \in \mathcal{N}_n} (\sum \limits_{\mathcal{Z}^{L}_{n,j}} \prod_{n_{k},n_h \in \mathcal{Z}^{L}_{n,j}} \frac {1} {\sqrt{d_k d_h}}) \cdot \mathbf{E}_{j}^{(\text{T})(0)},
\end{align}
\noindent where $\mathcal{Z}^{L}_{n,j}$ represents the maximum length $L$ of a possible path from the $n$-th region node and the $j$-th region node, with the intermediate connection nodes $k$ and $h$. The variables $d_k$ and $d_h$ refer to the degrees of these intermediate nodes. From the above Eq~\ref{eq:devitation}, it is important to note that the weight of $\mathbf{E}_{j}^{(\text{T})(0)}$ is non-learnable, which means it cannot be adjusted during the message passing process over noisy graph structures when generating the region embedding $\mathbf{E}_{n}^{(\text{T})(L)}$. $\mathcal{N}_n$ denotes the set of neighbour nodes of region $n$. 
Our framework provides a solution to this issue by introducing a learnable and adaptive knowledge distillation approach. Here, we analyze the gradients of our knowledge distillation with KL divergence alignment objective $\mathcal{L}^{(\text{KL})}$ given the corresponding embeddings $\tilde{\mathbf{E}}^{(\text{S})}_n$ of region $n$ via the student model as follows:

\begin{equation}
\begin{aligned}
\label{eq:pred_level}
&\frac {\partial \mathcal{L}^{(\text{KL})}}{\partial{\tilde{\textbf{E}}_{n}^{(\text{S})}}} 
=\sum \limits_{n =1}^{N} \sum \limits_{t'=T+1}^{T+H} \omega \cdot \frac{\partial(\tilde{\mathbf{E}}_{n,t'}^{(\text{T})},  \tilde{\mathbf{E}}_{n,t'}^{(\text{S})})}{\tilde{\mathbf{E}}_{n,t'}^{(\text{S})}},\\
&\omega = {\frac{1}{\text{softmax}(\tilde{\mathbf{E}}_{n,t'}^{(\text{S})})}}(-\frac{e^{\tilde{\mathbf{E}}_{j,t'}^{(\text{S})}}}{\sum\limits_{n'=1}^{N}e^{\tilde{\mathbf{E}}_{n',t'}^{(\text{S})}}})(-\frac{e^{\tilde{\mathbf{E}}_{n,t'}^{(\text{S})}}}{\sum\limits_{n'=1}^{N}e^{\tilde{\mathbf{E}}_{n',t'}^{(\text{S})}}}).
\end{aligned}
\end{equation}

\noindent Here, $\tilde{\mathbf{E}}_{n',t'}^{(\text{S})}$ represents the region embedding obtained from the TCN layers of the student model, denoted by the superscript $(S)$. The subscripts $n',t'$ indicate node $n'$ at time step $t'$. Similarly, $\tilde{\mathbf{E}}_{n,t'}^{(\text{T})}$ represents the region embedding of node $n$ at time step $t'$ obtained from the TCN layers of the teacher model. The derivations in Equation~\ref{eq:pred_level} demonstrate that the region embeddings are refined through the transferred knowledge from the teacher model using the derived weight $\omega$. While recursive message passing can lead to over-smoothing of the representations, our framework automatically adapts the knowledge transfer process, mitigating these over-smoothing effects.

\begin{table*}
\renewcommand\arraystretch{0.95}
\centering
\setlength{\abovecaptionskip}{0.2cm}
\setlength{\belowcaptionskip}{0.1cm}
\setlength{\tabcolsep}{0.001pt}
\footnotesize
\resizebox{\linewidth}{!}{
    \begin{tabular}{c|c c c| c c c |c c c| c c c| c c c}
        \toprule
        \multirow{2}*{Models}  & \multicolumn{3}{c|}{PeMS-Bay} & \multicolumn{3}{c|}{PeMSD4} &
        \multicolumn{3}{c|}{PeMSD8} & \multicolumn{3}{c|}{PeMSD3} & \multicolumn{3}{c}{PeMSD7}\\
        \cmidrule{2-16}
        &\multicolumn{1}{c}{MAE $\downarrow$} & \multicolumn{1}{c}{RMSE $\downarrow$} & \multicolumn{1}{c|}{MAPE $\downarrow$} & \multicolumn{1}{c}{MAE $\downarrow$} & \multicolumn{1}{c}{RMSE $\downarrow$} & \multicolumn{1}{c|}{MPAE $\downarrow$} & \multicolumn{1}{c}{MAE $\downarrow$} & \multicolumn{1}{c}{RMSE $\downarrow$} & \multicolumn{1}{c|}{MPAE $\downarrow$} & \multicolumn{1}{c}{MAE $\downarrow$} & \multicolumn{1}{c}{RMSE $\downarrow$} & \multicolumn{1}{c|}{MAPE $\downarrow$} & \multicolumn{1}{c}{MAE $\downarrow$} & \multicolumn{1}{c}{RMSE $\downarrow$} & \multicolumn{1}{c}{MAPE $\downarrow$}\\\midrule
        HA &2.88  &5.59  &6.82\% &38.03  &59.24  &27.88\%  &34.86  &52.04  &24.07\% &31.58  &52.39  &33.78\% &45.12  &65.64  &24.51\%\\ 
        VAR &2.32  &5.25  &5.61\% & 24.54 & 38.61 & 17.24\% & 19.19 & 29.80 & 13.10\% & 23.65 & 38.26 & 24.51\% & 50.22 & 75.63 & 32.22\%\\
        DSANet &2.16  &4.97  &5.54\%& 22.79 & 35.77 & 17.12\% & 17.14 & 26.96 & 11.32\% & 21.29 & 34.55 & 23.21\% & 31.36 & 49.11 & 14.43\%\\
        DCRNN &2.07  &4.74  &4.90\% & 24.70 & 38.12 & 14.17\% & 17.86 & 27.83 & 11.45\% & 17.99 & 30.31 & 18.34\% & 25.22 & 38.61 & 11.82\%\\
        STGCN &2.42  &5.33  &5.58\%& 22.70 & 35.55 & 14.59\% & 18.02 & 27.83 & 11.40\% & 17.55 & 30.42 & 17.34\% & 25.33 & 39.34 & 11.21\%\\
        GWN &1.95  &4.52  &4.63\%& 25.45 & 39.70 & 17.29\% & 19.13 & 31.05 & 12.68\% & 19.12 & 32.77 & 18.89\% & 26.39 & 41.50 & 11.97\%\\
       ASTGCN  &2.10  &4.77  &5.30\%& 22.93 & 35.22 & 16.56\% & 18.25 & 28.06 & 11.64\% & 17.34 & 29.56 & 17.21\% & 24.01 & 37.87 & 10.73\%\\
       LSGCN &2.13  &4.82  &5.18\%& 21.53 & 33.86 & 13.18\% & 17.73 & 26.76 & 11.30\% & 17.94 & 29.85 & 16.98\% & 27.31 & 41.16 & 11.98\%\\
       STSGCN &2.10  &4.74  &5.28\%& 21.19 & 33.65 & 13.90\% & 17.13 & 26.86 & 10.96\% & 17.48 & 29.21 & 16.78\% & 24.26 & 39.03 & 10.21\%\\
       AGCRN &1.96  &4.57  &4.69\%& 19.83 & 32.26 & 12.97\% & 15.95 & 25.22 & 10.09\% & 15.98 & 28.25 & 15.23\% & 22.37 & 36.55 & 9.12\%\\
       STFGNN &1.83  &4.33  &4.19\%& 19.83 & 31.88 & 13.02\% & 16.64 & 26.22 & 10.60\% & 16.77 & 28.34 & 16.30\% & 22.07 & 35.80 & 9.21\%\\
       STGODE &2.02  &4.40  &4.72\%& 20.84 & 32.82 & 13.77\% & 16.81 & 25.97 & 10.62\% & 16.50 & 27.84 & 16.69\% & 22.99 & 37.54 & 10.14\%\\
        Z-GCNETs &2.03  &4.38  &4.71\%& 19.50 & 31.61 & 12.78\% & 15.76 & 25.11 & 10.01\% & 16.64 & 28.15 & 16.39\% & 21.77 & 35.17 & 9.25\%\\
        TAMP &2.04  &4.45  &4.76\%& 19.74 & 31.74 & 13.22\% & 16.36 & 25.98 & 10.15\% & 16.46 & 28.44 & 15.37\% & 21.84 & 35.42 & 9.24\%\\
        FOGS &2.07  &4.51  &4.80\%& 19.74 & 31.66 & 13.05\% & 15.73 & 24.92 & 9.88\% & 15.89 & 25.74 & 15.13\% & 21.28 & 34.88 & \textbf{8.95\%}\\ 
        DSTAGNN &2.13  &4.79  &5.32\%& 19.30 & 31.46 &12.72\% & 15.67 & 24.77 & 9.94\% & 15.57 & 27.21 & 14.68\% & 21.42 & 34.51 & 9.01\%\\
        STKD &2.08  &4.56  &4.82\%  &19.86  &31.93 &13.18\%  &15.81  &25.07 &10.02\%  &16.03  &25.95 &15.76\%  &21.64  &34.96 &9.03\%\\
        KD-Pruning &2.23  &4.97  &5.34\%  &21.22  &34.63  &14.15\% &17.46  &27.09  &11.74\% &17.12  &29.87  &17.06\% &24.55  &38.17  &11.90\%\\
        \textbf{\model (Ours)} &\textbf{1.78}  &\textbf{3.88}  &\textbf{4.15\%}& \textbf{19.21} & \textbf{31.31} & \textbf{12.70\%} & \textbf{15.43} & \textbf{24.52} & \textbf{9.84\%} & \textbf{15.11} & \textbf{24.74} & \textbf{14.41\%} & \textbf{20.78} & \textbf{33.95} & \underline{8.98\%}\\
        \bottomrule
\end{tabular}}
\caption{Overall performance of traffic prediction on PeMS-Bay, PeMSD4, PeMSD8, PeMSD3 and PeMSD7}
\label{tab:overall_performance}
\end{table*}

%% file: eval.tex
\section{Evaluation}
\label{sec:eval}

\subsection{Experimental Setting}
\noindent \textbf{Datasets}. In this study, we conduct a series of experiments using real-life traffic flow datasets from California, specifically the PEMS3, PEMS4, PEMS7, PEMS8 and PeMS-Bay datasets released by~\cite{song2020spatial}. The traffic data is aggregated into 5-minute time intervals, resulting in 12 points of data per hour. Additionally, we construct the spatial graph of our traffic sensing regions based on the road network, which models the relationships between traffic flow patterns in different regions of the city. 

\noindent \textbf{Baselines}. We conduct a comprehensive evaluation of \model\ by comparing it against 18 baselines, including many state-of-the-art GNN-based traffic prediction models, which are categorized into six groups: 1) \textbf{GCN-based methods}: STGCN~\cite{STGCN},DCRNN~\cite{DCRNN}, GWN~\cite{shleifer2019incrementally}; 2) \textbf{GAT-based approaches}: ASTGCN~\cite{zhu2021ast}, LSTGCN~\cite{han2022lst}, DSTAGNN~\cite{lan2022dstagnn}; 3) \textbf{Differential GNNs}: STG-ODE~\cite{STGODE}; 4) \textbf{GNNs enhanced with Zigzag Persistence}: Z-GCNETs~\cite{chen2021z} and TAMP~\cite{chen2021tamp}; 5) \textbf{Hybrid spatio-temporal GNNs}: FOGS~\cite{rao2022fogs}, AGCRN~\cite{AGCRN}, STSGCN~\cite{song2020spatial}, STFGNN~\cite{li2021spatial}. 6) \textbf{Distillation methods}: KD-Pruning~\cite{izadi2024knowledge} and STKD~\cite{wang2024spatial}.

\subsection{Effectiveness Evaluation}
We evaluate the effectiveness of our method, \model, and the baselines on 5 datasets in terms of MAE, MAPE, and RMSE metrics, as shown in Table~\ref{tab:overall_performance}. Based on results, we have following observations:

\textbf{Superior Prediction Accuracy.} 
Our proposed method has consistently outperforms other baselines across all four datasets, in most evaluation cases. This can be attributed to several key factors that contribute to the effectiveness of our approach. \emph{Firstly}, we are able to successfully distill spatial and temporal dynamics from the teacher model. This enables the student to capture time-evolving traffic dependencies across geographical regions and time slots without relying on cumbersome message passing frameworks.  \emph{Secondly}, our adaptive knowledge distillation method, realized through our dual-level knowledge transfer, guides student learning with appropriate knowledge to alleviate the over-smoothing effects of the GNN architecture. \emph{Third}, by enabling cross-region and cross-time dependency modeling in an adaptive manner, our spatio-temporal knowledge distillation alleviates the effects of noisy adjacent relationships, contributing to the robustness of traffic flow prediction.

\noindent \textbf{Performance Comparison among Baselines}. 
Among the various baselines, we observe that methods such as STSGCN and STFGNN, which incorporate time-aware spatial dependency, perform better than approaches such as STGCN and DCRNN, which only consider stationary spatial correlations among regions. This highlights the importance of capturing temporal dynamics when encoding spatial dependency relationships among regions. \qr{In contrast to distillation methods like STKD and KD-pruning, which incorporate LSTM or GNNs into student models, potentially leading to over-smoothing and suboptimal performance, TCNs demonstrate superior efficacy in capturing temporal dynamics compared to LSTMs.}
Our proposed spatio-temporal knowledge distillation paradigm is designed to transfer time-aware spatial dependency knowledge from the teacher model to the MLP student model. By doing so, the student model is able to capture complex spatio-temporal patterns of traffic flow, resulting in state-of-the-art traffic prediction performance.

\subsection{Model Scalability Investigation}
To evaluate the efficiency of our proposed \model, we conduct experiments on the large PeMSD7 dataset competing with several state-of-the-art baselines. We conduct the experiments on a server with 10 cores of Intel(R) Core(TM) i9-9820X CPU @ 3.30GHz, 64.0GB RAM, and 4 Nvidia GeForce RTX 3090 GPU. The results for inference time and forecasting accuracy are shown in Table~\ref{tab:efficiency}. Our analysis yields two key observations. First, our \model\ achieves competitive performance in terms of accuracy metrics, \ie, MAE, MAPE, and RMSE. This is particularly noteworthy given the potential for over-smoothing on large-scale spatial region graphs, which our framework avoids by not explicitly introducing graph message passing and instead distilling denoised spatio-temporal knowledge into graph-less designations. Second, our \model\ achieves much faster inference time than the compared baseline models, which is attributed to the fact that \model\ does not require recursive graph-based information propagation operations during inference phase. While our traffic flow predictor is a simple graph-less neural network, its achieved superior performance suggests the effectiveness of our knowledge distillation paradigm in injecting complex global spatio-temporal dependencies across high-order region and time connections into the student. The ability to achieve high accuracy with fast inference time is particularly important in practical applications, where traffic forecasting models need to operate in real-time urban sensing.

\begin{table}[htb!]
\centering
\scriptsize
\begin{tabular}{cccccc}
\toprule
        Datasets& \multicolumn{5}{c}{PeMSD7} \\\midrule
Method  & \multicolumn{1}{c}{MAE $\downarrow$} & \multicolumn{1}{c}{RMSE $\downarrow$} & \multicolumn{1}{c}{MAPE $\downarrow$} & \multicolumn{1}{c}{Inference $\downarrow$} & \multicolumn{1}{c}{\qr{Faster x} \textbf{$\uparrow$}}  \\ \midrule
ASTGCN  & \multicolumn{1}{c}{24.01}    & \multicolumn{1}{c}{37.87}     & \multicolumn{1}{c}{10.73\%}     & \multicolumn{1}{c}{20.06s}                   & \multicolumn{1}{c}{\textbf{7.99} $\times$}                                            \\ 

STFGNN  & \multicolumn{1}{c}{22.07}    & \multicolumn{1}{c}{35.80}     & \multicolumn{1}{c}{9.21\%}     & \multicolumn{1}{c}{53.81s}                  & \multicolumn{1}{c}{\textbf{21.44} $\times$}                                           \\ 

STGODE  & \multicolumn{1}{c}{22.99}    & \multicolumn{1}{c}{37.54}     & \multicolumn{1}{c}{10.14\%}     & \multicolumn{1}{c}{24.79s}                   & \multicolumn{1}{c}{\textbf{9.88} $\times$}                  \\ 
DSTAGNN & \multicolumn{1}{c}{21.42}    & \multicolumn{1}{c}{34.51}     & \multicolumn{1}{c}{9.01\%}     & \multicolumn{1}{c}{110.06s}                   & \multicolumn{1}{c}{\textbf{43.85} $\times$}                                           \\
\textbf{Ours}   & \multicolumn{1}{c}{\textbf{20.78}}    & \multicolumn{1}{c}{\textbf{33.95}}     & \multicolumn{1}{c}{\underline{8.98\%}}     & \multicolumn{1}{c}{\textbf{2.51s}}          & \multicolumn{1}{c}{-}                                           \\ \midrule
        Datasets &\multicolumn{5}{c}{PeMS-Bay}\\\midrule
        Method  & \multicolumn{1}{c}{MAE $\downarrow$} & \multicolumn{1}{c}{RMSE $\downarrow$} & \multicolumn{1}{c}{MAPE $\downarrow$} & \multicolumn{1}{c}{Inference $\downarrow$} & \multicolumn{1}{c}{\qr{Faster x} \textbf{$\uparrow$}} \\ \midrule
        ASTGCN & \multicolumn{1}{c}{2.10}    & \multicolumn{1}{c}{4.77}     & \multicolumn{1}{c}{5.30}     & \multicolumn{1}{c}{40.08s}                   & \multicolumn{1}{c}{\textbf{7.82} $\times$}                                          \\ 
        STFGNN & \multicolumn{1}{c}{1.83}    & \multicolumn{1}{c}{4.33}     & \multicolumn{1}{c}{4.19\%}     & \multicolumn{1}{c}{98.18s}                  & \multicolumn{1}{c}{\textbf{19.18} $\times$}                                         \\ 
        STGODE & \multicolumn{1}{c}{2.02}    & \multicolumn{1}{c}{4.40}     & \multicolumn{1}{c}{4.72\%}     & \multicolumn{1}{c}{208.51s}                  & \multicolumn{1}{c}{\textbf{40.72} $\times$}                                     \\ 
        DSTAGNN &\multicolumn{1}{c}{2.13}    & \multicolumn{1}{c}{4.79}     & \multicolumn{1}{c}{5.32\%}     & \multicolumn{1}{c}{86.97s}               & \multicolumn{1}{c}{\textbf{16.99} $\times$}\\
         \textbf{Ours} & \multicolumn{1}{c}{\textbf{1.78}}    & \multicolumn{1}{c}{\textbf{3.88}}     & \multicolumn{1}{c}{\textbf{4.15\%}}     & \multicolumn{1}{c}{\textbf{5.12s}}               & \multicolumn{1}{c}{-}                       \\ \bottomrule   
\end{tabular}
\caption{Model Efficiency Study}
\label{tab:efficiency}
\end{table}

\subsection{Ablation Study and Effectiveness Analyses}
\begin{figure}
\centering
\begin{tabular}{c c}
  \begin{minipage}{0.20\textwidth}
  \includegraphics[width=\textwidth]{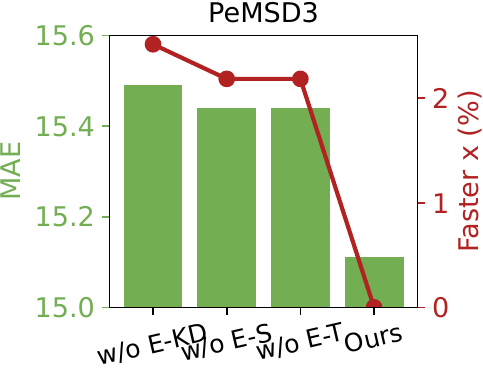}
  \end{minipage}
  &
  \begin{minipage}{0.20\textwidth}
    \includegraphics[width=\textwidth]{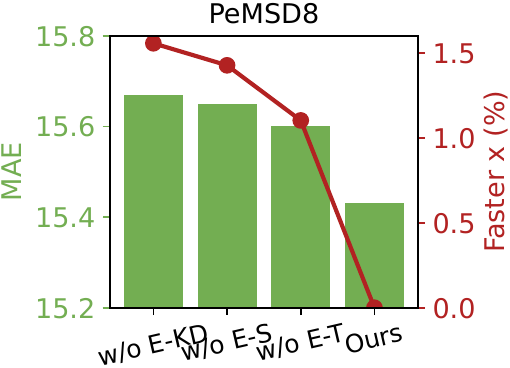}
  \end{minipage}
 \end{tabular}
\caption{Ablation study of sub-modules in our spatio-temporal knowledge distillation paradigm.}
\label{fig:ablation}
\end{figure}

To assess the impact of each component in our knowledge distillation framework on prediction results and speed, we conducted an ablation study across four traffic datasets using model variants. These variants include: 1) \emph{w/o E-KD}, which disables embedding-level knowledge distillation for transferring spatio-temporal signals from the latent representation space; 2) \emph{w/o E-S}, which omits adaptive embedding alignment with spatial information by removing $\mathcal{L}^{(\text{P})}$ from the joint loss $\mathcal{L}$; and 3) \emph{w/o E-T}, which excludes $\mathcal{L}^{(\text{E})}$ from $\mathcal{L}$ to avoid capturing temporal information during embedding-level knowledge distillation. The results, presented in Figure~\ref{fig:ablation}, reveal key observations. Firstly, the \emph{w/o E-KD} variant performs significantly worse than our full model, highlighting the crucial role of embedding-level knowledge distillation in transferring spatio-temporal signals. Notably, KL divergence is the most computationally demanding component of our model. Secondly, the superior performance of our model compared to \emph{w/o E-S} and \emph{w/o E-T} demonstrates the effectiveness of adaptive embedding alignment across both spatial and temporal domains in capturing complex cross-location and cross-time traffic dependencies.

\subsection{Hyperparameter Study}
\begin{figure}
\centering
\begin{tabular}{c c }
  \begin{minipage}{0.20\textwidth}
	\includegraphics[width=\textwidth]{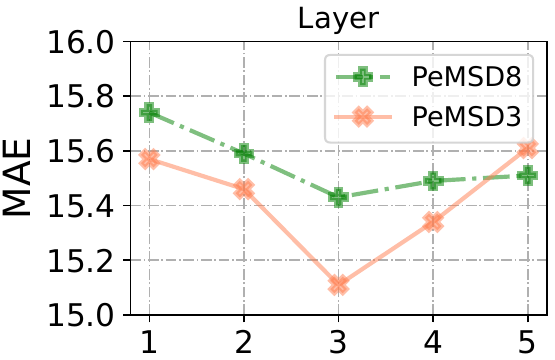}
  \end{minipage}
  &
  \begin{minipage}{0.20\textwidth}
    \includegraphics[width=\textwidth]{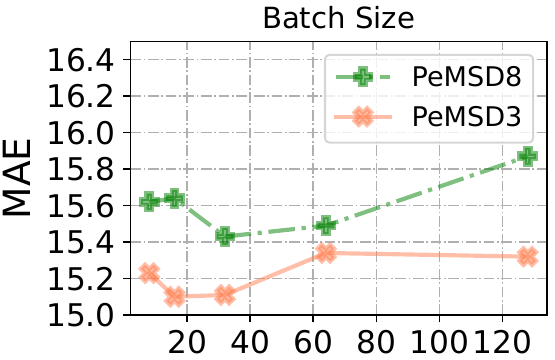}
  \end{minipage}
\end{tabular}
\caption{Hyparameter study on PeMSD8 and PeMSD3 in terms of MAE.}
\label{fig:hyperparameter}
\end{figure}

The aim of this section is to evaluate the effects of key hyperparameters on the performance of our framework, \model. We present our results on PeMSD8 and PeMSD3 datasets in terms of MAE and RMSE in Figure~\ref{fig:hyperparameter}. We summarie our observations as follows: 1) Figure ~\ref{fig:hyperparameter} show the effect of the number of MLP layers (ranging from $\left\{1,2,3,4,5\right\}$) and varying batch size (ranging from $\left\{2^3, 2^4, 2^5, 2^6, 2^7\right\}$) on performance. Our framework, \model, achieves the best performance on PeMSD8 and PeMSD3 when the number of layers is 3 and the batch size is 32. Even when \model\ achieves the worst performance, it still outperforms most of the baselines. These results suggest that the performance of our \model\ is not sensitive to the MLP depth and batch size. 2) $\lambda_1, \lambda_2$ serve as loss weights to control how strongly our prediction-level and embedding-level restrict the joint model training. Figure ~\ref{fig:hyperparameter} show that $\lambda_1$ and $\lambda_2$ jointly affect the strength of the optimization of knowledge distillation. We find that a larger weight of distillation causes performance maintenance, enabling MLP to learn sufficient knowledge.

%% file: relate.tex
\section{Related Work}
\label{sec:relate}
\noindent \textbf{Traffic Flow Prediction}. 
Numerous neural network architectures have been proposed for traffic prediction, including convolutional neural networks (CNNs) ~\cite{zhang2017deep}, recurrent neural networks (RNNs)~\cite{lv2018lc}, attention mechanisms~\cite{yao2019revisiting}, and graph neural networks (GNNs)~\cite{li2021spatial}. CNNs have proven effective in modeling citywide traffic maps as images for spatio-temporal pattern encoding~\cite{diao2019dynamic,zhang2017deep}, while RNNs excel at capturing temporal dependencies in time-evolving traffic data~\cite{lv2018lc}. To model spatial traffic similarities adaptively, research has explored spatial dependency graphs with learnable region adjacency matrices~\cite{shleifer2019incrementally,AGCRN,rao2022fogs,lan2022dstagnn}. For instance, DSTAGNN~\cite{lan2022dstagnn} utilizes a multi-head attention mechanism to exploit spatial correlations with multi-scale neighborhoods. FOGS~\cite{rao2022fogs} learns a spatial-temporal correlation graph using first-order gradients during training. While GNN-enhanced traffic prediction models hold promise, their computational complexity hinders scalability and real-world deployment. This study addresses this challenge by leveraging spatio-temporal knowledge distillation to reduce inference time, enabling our model to effectively scale to larger datasets.

Recent traffic prediction methods have employed Multilayer Perceptron (MLP)-based networks, including STID~\cite{shao2022spatial} and ST-MLP~\cite{wang2023st}.  STID leverages MLPs to capture spatio-temporal dynamics using spatial and temporal identity mappings.  ST-MLP~\cite{wang2023st} focuses on using MLPs exclusively, relying on predefined spatial graph structures to improve efficiency.  While Transformer-based approaches, such as STAEformer~\cite{liu2023spatio}, show promising performance, their high computational cost and resource requirements hinder their efficiency and scalability.

%% file: conclusion.tex
\section{Conclusion}
\label{sec:conclusoin}
We propose \model, a scalable and high-performance traffic flow prediction framework  that offers both model efficiency and generalization capability, which are often lacking in existing solutions. We draw inspiration from the knowledge distillation paradigm to achieve efficiency and retain the awareness of high-order spatial-temporal traffic dependencies across locations and time. To this end, we perform both explicit prediction-level and implicit embedding-level distillation to transfer spatio-temporal knowledge from a cumbersome GNN teacher to a simple yet effective MLP student. Additionally, our model adopts a new adaptive model alignment schema to further enhance the student model by alleviating over-smoothing effects. In future work, we plan to enhance our spatio-temporal knowledge distillation with causal inference, to identify confounding variables that may affect the spatio-temporal data and adjust for their effects in the knowledge distillation process.

%% file: appendix.tex
\appendix \section{Appendices}
\label{sec:appendix}

\subsection{Algorithm of \model}
\begin{algorithm}[h]
    \caption{The \model\ Learning Algorithm}
    \label{alg:overall}
    \KwIn{
        Historical observation tensor $\mathbf{X} \in \mathbb{R}^{N \times T}$ and the spatial graph $\mathcal{G}$, $\tau_2$, $\tau_3$, $\lambda_1$,$\lambda_2$, learning rate $\eta$, maximum training epochs $S$ 
    }
    \KwOut{
        trained parameters in $\Theta$ 
    }
    Initialize all parameters in $\Theta$;\\
    Train teacher model by Equation~\ref{eq:lossgcn}, and collect the traffic embeddings at each time interval $t$, denoted as $E^{(\text{T})}_{t}$\\
    \For{$epoch = 1, 2,..., S$}{
        Calculate the loss $\mathcal{L}^{(\text{S})}$ by Equation~\ref{eq:lossgcn};\\
        Perform prediction-level distillation and compute the KL loss $\mathcal{L}^{(\text{KL})}$ by Equation~\ref{eq:losspredis};\\
        Perform embedding-level distillation from spatial and temporal dimensions by Equation~\ref{eq:loss_sp};\\
        Calculate the overall loss $\mathcal{L}$ by Equation~\ref{eq:loss_sum};\\
        \For{Optimizing parameters on weight factor in prediction-level}{
        Calculating weight factor according to Eq~\ref{eq:pred_level}}
        \For{$\theta \in \Theta$}{
        $\theta = \theta-\eta \cdot \frac{\partial \mathcal{L}^{(\text{S})}}{\partial \theta}$
        }
    }
\textbf{Return} all parameters $\Theta$ 
\end{algorithm}

\begin{table}[t]
\renewcommand\arraystretch{1.0}
\centering
\setlength{\abovecaptionskip}{0.2cm}
\setlength{\belowcaptionskip}{0.1cm}
\setlength{\tabcolsep}{3pt}
\footnotesize

\resizebox{\linewidth}{!}{
\begin{tabular}{ccccc}
\toprule
Datasets & \#Sensors &Time Period & Time Steps & Interval \\ \midrule
PeMSD4   &307         &2018/1/1-2018/2/28       &16,992            &5 minutes               \\
PeMSD8   &170         &2016/7/1-2016/8/31       &17,856            &5 minutes               \\
PeMSD3   &358         &2018/9/1-201811/30       & 26,208           &5 minutes               \\
PeMSD7   &883         &2017/5/1-2017/8/31       &28,224            &5 minutes               \\ 
PeMS-Bay   &325         &2017/1/1-2017/5/31       &52,116           &5 minutes               \\\bottomrule
\end{tabular}
}
\caption{Data Description and Statistics.}
\label{tab:data_sta}
\end{table}

\subsubsection{More Related Work}
\noindent \textbf{Knowledge Distillation for Graphs}. 
Knowledge distillation on graphs provides a promising approach for transferring knowledge from complex teacher GNNs to smaller student models, effectively reducing computational costs while preserving accuracy~\cite{guo2022linkless,wu2022knowledge,feng2022freekd,qin2021slow}. This technique has been applied to various graph applications, including node/graph classification~\cite{zhang2021graph,he2022compressing}, 2022], social media analysis~\cite{qian2021distilling}, and recommender systems~\cite{tao2022revisiting}. Adversarial training, involving a discriminator and generator, has been employed to enhance knowledge distillation~\cite{he2022compressing}. ~\citet{qian2021distilling} applied knowledge distillation to a heterogeneous graph for analyzing drug trafficking from social media data. ~\citet{tao2022revisiting} proposed a distillation-enhanced relational encoder to improve recommendation accuracy by capturing user-item interactions and social connections. STKD~\cite{wang2024spatial} is a method that adopts distillation via 1D CNN and LSTM to detect traffic anomalies. However, 1D CNNs are limited by fixed-size windows, hampering their ability to capture long-range dependencies crucial for modeling complex temporal patterns in traffic data.

\section{Acknowledgements}
\label{sec:ack}
This project is supported by HKU-SCF FinTech Academy and Shenzhen-Hong Kong-Macao Science and Technology Plan Project (Category C Project: SGDX202108 23103537030) and Theme-based Research Scheme T35-710/20-R. The Australian Research Council partially supports this work under the streams of Future Fellowship (Grant No. FT210100624), and the Linkage Project (Grant No. LP230200892). We also appreciate the help of Dr.Huang, Dr.Xia, Dr.Li and Dr.Tang.